\documentclass[11pt]{article}

\usepackage{acl}

\usepackage{amsmath}
\usepackage{algorithm}
\usepackage{times}
\usepackage{latexsym}
\usepackage{booktabs}
\usepackage[T1]{fontenc}

\usepackage[utf8]{inputenc}
\usepackage{xcolor}
\usepackage{array}
\usepackage{graphicx} 
\usepackage{booktabs}
\usepackage{tabularx}
\usepackage{multirow}
\usepackage{microtype}
\usepackage{arydshln}
\usepackage{inconsolata}

\usepackage{graphicx}

\title{DE-NER : Zero-shot Named Entity Recognition via Dialogue Elicitation of Large Language Models}

\author{Xuankang Zhang\\
  \textit{Yunnan University} \\
  \texttt{zhangxuankang@stu.ynu.edu.cn} \\\And
  Jiangming Liu\thanks{*Corresponding author.}\\
  \textit{Yunnan University} \\
  \texttt{jiangmingliu@ynu.edu.cn} \\}

\begin{document}

\maketitle
\begin{abstract}
Recent advancements of zero-shot Named Entity Recognition (NER) establish strong baselines by formulating sequence labeling into question answering where Large Language Models (LLMs) can be naturally adopted.
However, existing LLM-based zero-shot NER methods suffer from the limitations of prompt and demonstration engineering. To address these issues with minimal human interventions, we introduce DE-NER, a dialogue elicitation framework which elicits the chatting ability of LLMs to fully extract the knowledge encoded in LLMs. Our experiments demonstrate that the proposed method outperform the competitive baselines in zero-shot settings across multiple benchmarks, with an average improvement of 3.75\% F1 points. Codes are released in \url{https://github.com/kkkenshi/DE-NER}.
\end{abstract}

\section{Introduction}

Large Language Models (LLMs) demonstrate remarkable zero-shot generalization across a wide range of NLP tasks \cite{Brown2020LanguageMA}. Recent advancements of zero-shot Named Entity Recognition (NER) transform the traditional sequence labeling into the tasks where LLMs can be naturally adopted, such as question-answering \cite{wei2023chatie}, template generation \cite{Cui2021TemplateBasedNE}, and instruction designing \cite{wang2023instructuie}. However, existing methods rely on labor-intensive prompt engineering or in-context learning with demonstrations that are sensitive to human design. 

Recent research in broader NLP domains shift towards dialogue-based paradigms \cite{liang2024encouraging,Li2023CAMELCA}. By allowing models to engage in multi-turn interactions, these frameworks can clarify implicit information and integrate context effectively \cite{shinn2023reflexion}. \citet{Andukuri2024STaRGATETL} show that teaching models to ask clarifying questions significantly improves their ability to elicit latent constraints and resolve ambiguity in reasoning tasks. 
The interactive capability of LLMs holds the key to addressing the precision and boundary challenges in zero-shot NER.

\begin{figure}[!tp]
    \centering
    \includegraphics[width=\columnwidth]{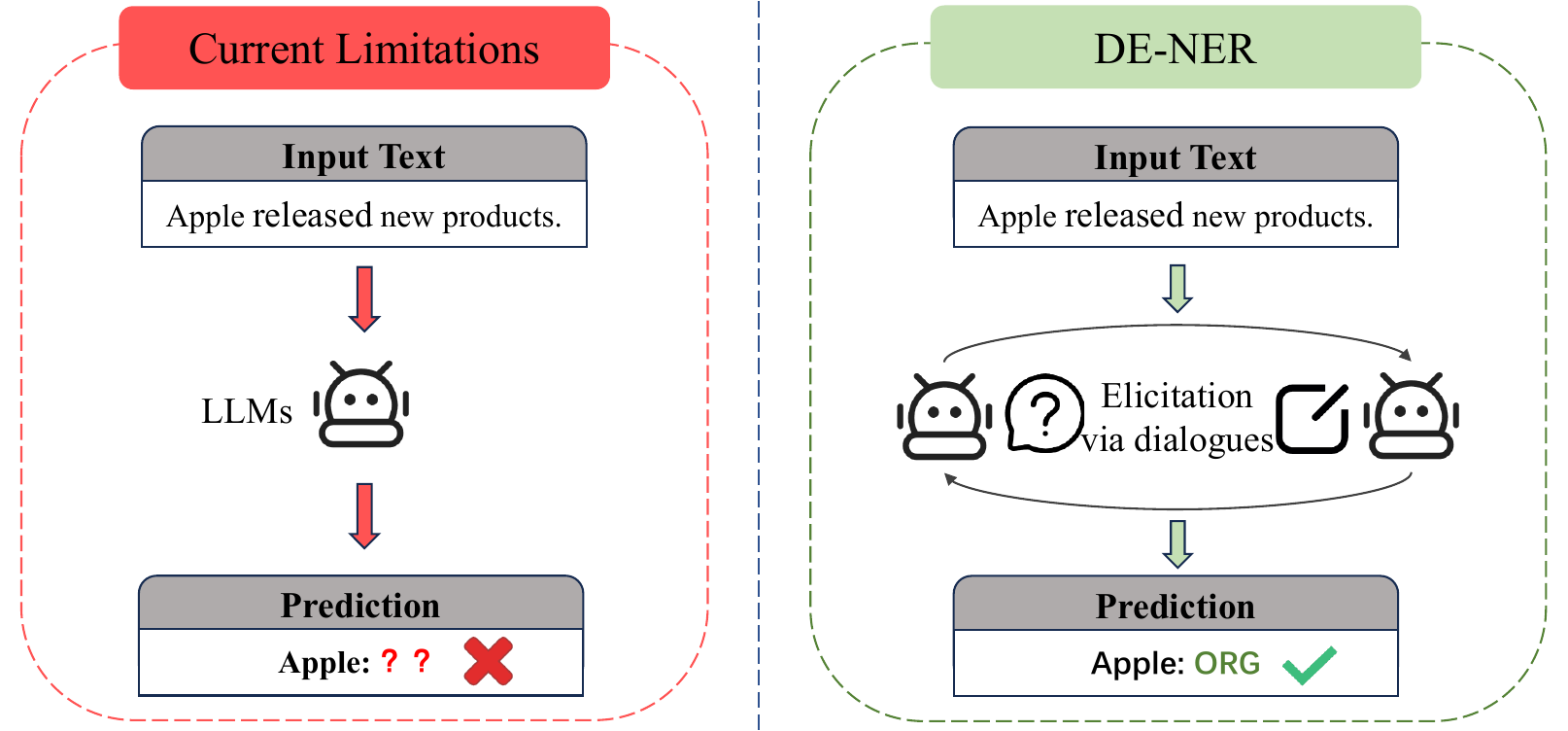}
    \caption{Conventional NER methods and our proposed DE-NER framework.}
    \label{fig:introduction}
\end{figure}

Motivated by the success of interaction with LLMs, we explore whether LLMs can improve their own ability to extract entities by engaging in dialogues with LLMs. To this end, we introduce \textbf{DE-NER}, a \textbf{D}ialogue \textbf{E}licitation of LLMs for zero-shot \textbf{NER}. As shown in Figure \ref{fig:introduction}, previous single-turn paradigms struggle with high uncertainty, while our framework elicits the LLMs by self-play to clarify the implicit boundary information that is required by NER. For each iteration, a questioner generates questions that can be answered by a roleplayer, and a final responser model integrates the entire dialogue to refine NER predictions without any human engagement. This process automatically elicits the model to reduce ambiguity in entity boundaries without relying on external knowledge bases or manual annotations.

We evaluate our method on CoNLL03, WikiGold, and GENIA in zero-shot settings. Our proposed models demonstrate that the strong NER ability for challenging domain-specific entities by outperforming the conventional zero-shot prompting methods. The main contributions are summarized as follows:
\begin{itemize}
    \item We present a novel Dialogue Elicitation of LLMs for zero-shot NER without depending on external retrieval or knowledge bases.
    
    \item We propose an efficient dialogue trajectory optimization method to enable DE-NER ask clarifying elicited questions to LLMs that can response correctly for NER.
    
    \item Experimental results show that the proposed model outperforms the competitive baselines across benchmarks in zero-shot settings.
\end{itemize}

\begin{figure*}[!tp]
    \centering
    \includegraphics[width=\textwidth]{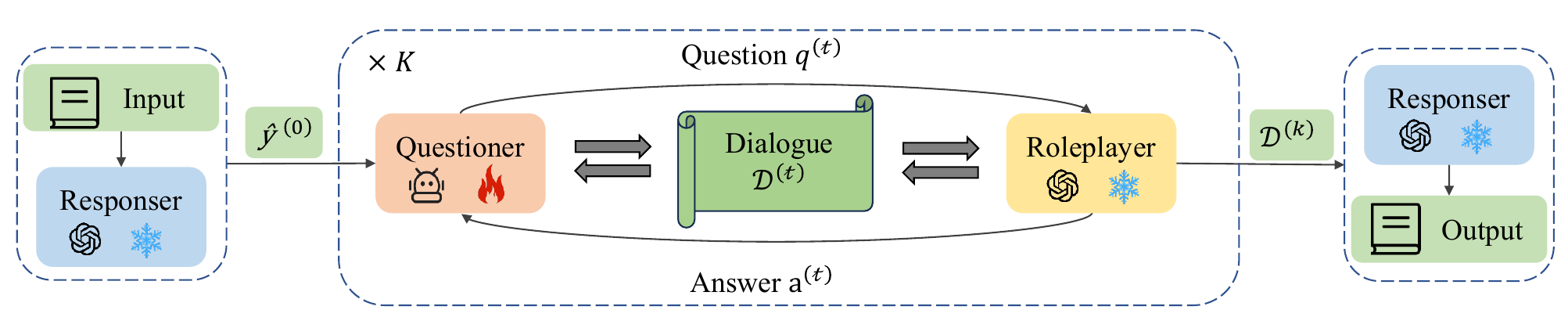} 
    \caption{Framework of DE-NER.}
    \label{fig:framework}
\end{figure*}

\section{Related Work}

\subsection{Zero-Shot Named Entity Recognition} 

\paragraph{Prompt-Based}
 Early LLM-based methods formulate NER as a template-based generation task \cite{Cui2021TemplateBasedNE,ding-etal-2022-prompt} or a question-answering task \cite{wei2023chatie}, converting label descriptions into natural language queries to achieve extraction. More recent works fine-tune LLMs on diverse task collections to explicitly follow zero-shot extraction instructions \cite{wang2023instructuie}. However, they suffer from prompt sensitivity where minor variations in instruction wording can lead to unstable predictions.

\paragraph{Self-annotated Demonstrations} 
A recent trend involves converting zero-shot tasks into dynamic few-shot settings by generating demonstrations. ReverseNER \cite{wang2024reversener} constructs a reliable example library by reversing the generation process from entity definitions, while LLMaAA \cite{zhang2023llmaaa} utilizes LLMs as active annotators to synthesize pseudo-labels. Although these generate-then-retrieve pipelines alleviate the lack of supervision, they incur computational overhead due to the requirement for offline data synthesis.

\paragraph{Knowledge-Enhanced}
This line of work enhances zero-shot NER by incorporating external knowledge, such as label semantics, ontology descriptions, or structured resources, to support entity grounding and type disambiguation \cite{liang2020bond,jin2024genegpt,cocchieri-etal-2025-zeroner}. 
These methods leverage descriptive knowledge rather than labeled instances, enabling annotation-free inference. 
Instead, our approach resolves ambiguity through multi-turn dialogues without accessing external knowledge sources.

\subsection{Interactive Reasoning with LLMs}

\paragraph{Chain of Thought Reasoning}

CoT encourages LLMs to generate intermediate rationale before final prediction. This paradigm adapts to NER to enhance contextual understanding and few-shot generalization \cite{wei2022chain}. Recent works \cite{xu2023multi} propose multi-task instruction frameworks that integrate reasoning steps to guide entity extraction. \citet{li2024rt} combine retrieval mechanisms with chain-of-thought reasoning to boost performance on complex medical entities. However, the methods generally operate as a static generation process without any feedback from users, leading the challenge of error accumulation.

\paragraph{Agent Interaction}
Recent studies extend interactive LLMs to agent-based reasoning, self-refinement, and multi-agent collaboration \cite{shinn2023reflexion,liang2024encouraging}. This paradigm has recently been explored in information extraction, where agents are used for cross-agent debate in zero-shot extraction \cite{lu2025crossagentie} or for document-level extraction with external tools \cite{li2025aid}. Different from these post-refinement frameworks, DE-NER localizes agent interaction to entity-level clarification, where a trainable Questioner proactively asks about type ambiguity, missed entities, and type errors before final prediction


\section{DE-NER}
We propose a novel model for zero-shot NER that elicits LLMs to accomplish the task with self-play. As shown in Figure~\ref{fig:framework}, the proposed DE-NER consists of three functional roles implemented by large language models: a Questioner ($\textit{Q}$), a Roleplayer ($\textit{P}$) and a Responser ($\textit{R}$). The Questioner is responsible for raising clarifying questions; the Roleplayer provides answers based on the dialogue history. After several rounds of question answering, the Responser outputs the final predictions. The self-play pipeline of question and answer is without any human interactions.


\subsection{Interaction with LLMs}
\label{Inference}

Given an input sentence
$\mathbf{x} = \{x_1, \ldots, x_n\},$
the goal of NER is to predict a corresponding sequence of entities
$
\hat{\mathbf{y}} = \{y_1, \ldots, y_m\}.
$
We formulate zero-shot NER as a dialogue-driven inference process.
The Responser first produces an initial zero-shot NER prediction without any dialogue context
$
\hat{\mathbf{y}}^{(0)} = \mathrm{\textit{R}}(\mathbf{x}).
$
This preliminary prediction serves as a hypothesis that will be examined and refined through interaction. The dialogue history is initialized as an empty set,
$
\mathcal{D}^{(0)} = \emptyset.
$

For each dialogue turn $t \in \{1, \ldots, N\}$, the Questioner generates a clarification question by conditioning on the initial prediction and the dialogue history accumulated so far:
$
q^{(t)} = \mathrm{\textit{Q}}\big(\hat{\mathbf{y}}^{(0)}, \mathcal{D}^{(t-1)};\theta\big).
$
The Roleplayer answers each clarification question by grounding its response strictly in the input sentence:
$
a^{(t)} =\textit{P}\big(q^{(t)}, \mathcal{D}^{(t-1)}\big).
$
 Each question–answer pair $(q^{(t)}, a^{(t)})$ is appended to the dialogue history:
$
\mathcal{D}^{(t)} = \mathcal{D}^{(t-1)} \cup \{(q^{(t)}, a^{(t)})\}.
$

Given $N$ clarification turns of interactions, the Responser produces the final NER prediction by conditioning on both the input sentence and the full dialogue history:
$
\hat{\mathbf{y}}^{(N)} = \textit{R}(\mathbf{x}, \mathcal{D}^{(N)}).
$
Through this structured interaction process, the model is able to surface implicit constraints in the input text, revise earlier decisions, and correct entity boundaries or types before producing the final zero-shot NER output.


\begin{table}[!t] 
    \centering
    \renewcommand{\arraystretch}{1.1} 
    
    \resizebox{\linewidth}{!}{
    \begin{tabular}{lcccccc}
        \toprule
        \textbf{Model} & $N$ & \textbf{LOC.} & \textbf{ORG.} & \textbf{PER.} & \textbf{MISC} & \textbf{AVG.} \\
        \midrule
        
        \multicolumn{7}{c}{\textbf{CoNLL03}} \\ 
        \midrule
        Prompt & -- & 72.08$\pm$0.82 & 58.92$\pm$2.19 & 91.15$\pm$1.63 & 21.17$\pm$7.34 & 70.58$\pm$1.29 \\
        \midrule
        \multirow{3}{*}{Base} & 2  & 71.34$\pm$0.76 & 61.35$\pm$1.77 & 93.60$\pm$0.32 & 42.21$\pm$7.88 & 72.28$\pm$0.94 \\
                              & 3  & 71.52$\pm$0.67 & 61.65$\pm$1.37 & 93.92$\pm$0.30 & 22.22$\pm$6.05 & 70.78$\pm$0.20 \\
                              & 4  & 70.08$\pm$0.87 & 62.32$\pm$1.19 & 93.44$\pm$0.22 & 19.35$\pm$4.07 & 70.72$\pm$0.35 \\
        \hdashline
        \multirow{3}{*}{DE-NER} & 2  & \textbf{72.86$\pm$0.22} & 63.08$\pm$2.39 & 93.69$\pm$0.78 & \textbf{49.59$\pm$3.87} & 73.96$\pm$0.92 \\
                                & 3  & 72.55$\pm$0.26 & 64.11$\pm$2.33 & 93.73$\pm$0.52 & 44.82$\pm$8.04 & 73.85$\pm$0.99 \\
                                & 4  & 71.87$\pm$0.16 & \textbf{65.08$\pm$2.27} & \textbf{94.20$\pm$0.41} & 49.12$\pm$0.81 & \textbf{74.35$\pm$0.39} \\
        
        \midrule
        \multicolumn{7}{c}{\textbf{WikiGold}} \\ 
        \midrule
        Prompt & -- & 80.34$\pm$1.36 & 69.53$\pm$1.17 & 91.72$\pm$1.23 & 36.14$\pm$1.54 & 71.55$\pm$1.14 \\
        \midrule
        \multirow{3}{*}{Base} & 2  & 80.93$\pm$0.06 & \textbf{71.95$\pm$1.42} & 92.09$\pm$1.21 & 42.90$\pm$6.15 & 73.53$\pm$1.26 \\
                              & 3  & 79.79$\pm$1.46 & 70.72$\pm$0.27 & 91.86$\pm$0.50 & 40.10$\pm$1.40 & 72.43$\pm$0.59 \\
                              & 4  & 80.18$\pm$1.44 & 69.11$\pm$0.82 & 91.62$\pm$1.27 & 40.73$\pm$2.99 & 71.98$\pm$0.89 \\
        \hdashline
        \multirow{3}{*}{DE-NER} & 2  & \textbf{82.76$\pm$0.89} & 71.65$\pm$1.15 & 92.28$\pm$0.52 & \textbf{49.89$\pm$5.11} & \textbf{74.97$\pm$1.26} \\
                                & 3  & 81.56$\pm$0.96 & 70.58$\pm$1.03 & 91.88$\pm$0.61 & 48.35$\pm$3.49 & 74.04$\pm$0.91 \\
                                & 4  & 81.17$\pm$0.76 & 70.52$\pm$2.09 & \textbf{92.32$\pm$0.61} & 46.64$\pm$5.30 & 73.53$\pm$1.41 \\
        \bottomrule
    \end{tabular}
    }
    \vspace{2pt} 
    \resizebox{\linewidth}{!}{
    \begin{tabular}{lccccccc}
        \textbf{Model} & $N$ & \textbf{DNA} & \textbf{RNA} & \textbf{LINE.} & \textbf{TYPE.} & \textbf{PROT.} & \textbf{AVG.} \\
        \midrule
        \multicolumn{8}{c}{\textbf{GENIA}} \\
        \midrule
        Prompt & -- & 21.36 & 46.15 & 43.94 & 51.35 & 54.80 & 47.65 \\
        \midrule
        \multirow{3}{*}{Base} & 2  & \textbf{33.62$\pm$2.09} & 48.08$\pm$2.72 & \textbf{45.55$\pm$0.35} & 51.93$\pm$2.44 & 57.97$\pm$0.25 & 51.48$\pm$0.09 \\
                              & 3  & 30.16$\pm$2.60 & 45.64$\pm$2.45 & 42.98$\pm$1.58 & 48.69$\pm$0.70 & 59.03$\pm$2.68 & 50.61$\pm$1.06 \\
                              & 4  & 29.95$\pm$2.85 & 46.89$\pm$2.67 & 44.19$\pm$0.35 & 49.26$\pm$0.16 & 57.43$\pm$0.74 & 50.02$\pm$0.13 \\
        \hdashline
        \multirow{3}{*}{DE-NER} & 2  & 28.78$\pm$1.41 & 49.33$\pm$2.76 & 44.80$\pm$1.22 & \textbf{53.31$\pm$1.21} & \textbf{59.16$\pm$0.95} & \textbf{51.70$\pm$0.01} \\
                                & 3  & 27.52$\pm$0.20 & 45.30$\pm$2.92 & 44.86$\pm$2.34 & 50.95$\pm$0.75 & 58.62$\pm$0.23 & 50.68$\pm$0.18 \\
                                & 4  & 25.33$\pm$1.62 & \textbf{51.05$\pm$12.65} & 42.21$\pm$1.40 & 49.91$\pm$1.34 & 58.68$\pm$0.49 & 50.09$\pm$0.08 \\
        \bottomrule
    \end{tabular}
    }
    \caption{Results(\%) with GPT-3.5-Turbo on CoNLL03, WikiGold , GENIA respectively. The best scores are \textbf{bold}.}
    \label{tab:overall_results}
\end{table}

\subsection{Training with Dialogue Trajectory}
\label{generation}

To enhance the Questioner ability to ask effective questions under a zero-shot setting, we propose a self-training strategy that constructs supervision signals from dialogue trajectories. Given an unlabeled corpus, we sample a set of sentences $\{\mathbf{x}_i\}$. 
For each sentence, we independently run the dialogue-based inference procedure $K$ times with a high temperature to encourage diverse questioning strategies:
$
\mathcal{T}_{i} = \{(\mathcal{D}^{(N)}_{i,1}, \hat{\mathbf{y}}^{(N)}_{i,1}), \ldots, (\mathcal{D}^{(N)}_{i,K}, \hat{\mathbf{y}}^{(N)}_{i,K})\},
$
where $\mathcal{D}^{(N)}_{i,k}$ represents the sequence of question-answer pairs in the $k$-th run, and $\hat{\mathbf{y}}^{(N)}_{i,k}$ is the resulting entity prediction.

We derive a high-confidence pseudo-label $\tilde{\mathbf{y}}_i$ via entity-level majority voting:
$
\tilde{\mathbf{y}}_i = \mathrm{Vote}\big(\{\hat{\mathbf{y}}^{(N)}_{i,k}\}_{k=1}^{K}\big)$,
where $\mathrm{Vote}(\cdot)$ denotes an entity span and its associated type appear in more than half of the $K$ predictions.

Finally, we identify the optimal dialogue trajectory $\mathcal{D}^*_i$ by maximizing the overlap with the pseudo-label:
$\mathcal{D}^*_i = \arg\max_{\mathcal{D}^{(N)}_{i,k} \in \mathcal{T}_i} \mathrm{\text{Overlap}}(\hat{\mathbf{y}}^{(N)}_{i,k}, \tilde{\mathbf{y}}_i)$.
The Questioner is optimized according to $\mathcal{D}_i^{*} = \{(q^{(1)}, a^{(1)}), \ldots, (q^{(N)}, a^{(N)})\}$, while the Roleplayer and the Responser remain frozen. The training loss minimizes the negative log-likelihood of generating the clarification question $q^{(t)}$ at each dialogue turn $t \in (1,N)$:
$
\mathcal{L}_{\text{train}}
= - \frac{1}{N}
\sum_{t=1}^{N}
\log p_{\theta}\big(q^{(t)} \mid \mathbf{x}, \mathcal{D}^{(t-1)}\big),
$
where $p_{\theta}(\cdot)$ is the probability of generating question of $q^{(t)}$ by the questioner \textit{Q}.


\section{Experiments}

We carry out experiments on zero-shot standard benchmarks of CoNLL03~\cite{sang2003introduction}, WikiGold~\cite{balasuriya2009named} and GENIA~\cite{ohta2002genia} spanning general and biomedical domains. 

\subsection{Baselines and Settings}
All baselines utilize the same base model for entity prediction to ensure fair comparison. \textbf{Prompt} is instructed to extract entities directly from the input text without any dialogue interaction or external examples.
\textbf{Base} adopts a questioner to generate
multi-turn questions without any self-training.
\textbf{DE-NER} is our proposed model where the questioner is optimized using the dialogue trajectories selected via the voting process.





We use Mistral-7B-Instruct-v0.2 as the questioner, while both the Roleplayer and the Responser are implemented using GPT-3.5-Turbo. We take $N \in \{2, 3, 4\}$ rounds of clarification dialogue. The temperature is 0.8 for questioners and 0.2 for roleplayer and responser with 3 runs.\footnote{Detailed prompts can be found in Appendix~\ref{appendix}.}




\begin{table}[t]
    \centering
    \renewcommand{\arraystretch}{1}
    \resizebox{\linewidth}{!}{
    \begin{tabular}{lccccc}
        \toprule
        & Prompt & Base & DE-NER \\
        \midrule
        Mistral-7B-instruct-v0.2 &36.85 &48.59 &25.56 \\
        GPT-3.5-Turbo &70.58 &72.28 &73.96 \\
        GPT-4o-mini &75.54 & 75.81 &76.21 \\
        \bottomrule
    \end{tabular}
    }
    \caption{Results(\%) on CoNLL03 of the models for various LLMs.}
    \label{tab:mistral_results}
\end{table}


\subsection{Results}
\paragraph{Main Result}
Tables~\ref{tab:overall_results} show the results on CoNLL03, WikiGold, and GENIA, respectively. Base achieves consistent gains over the Prompt, validating the effectiveness of multi-turn interaction while DE-NER yields the best performance by obtaining 74.35, 74.97 and 51.70 F1 scores on CoNLL03, WikiGold and GENIA, respectively. 

\begin{table}[!t]
    \centering
    \renewcommand{\arraystretch}{1.1}
    \resizebox{\linewidth}{!}{
    \begin{tabular}{lccccccc}
        \toprule
        \textbf{Model} & $N$ & $k$ & \textbf{LOC.} & \textbf{ORG.} & \textbf{PER.} & \textbf{MISC} & \textbf{AVG.} \\
        \midrule
        Prompt & -- & -- & 72.08$\pm$0.82 & 58.92$\pm$2.19 & 91.15$\pm$1.63 & 21.17$\pm$7.34 & 70.58$\pm$1.29 \\
        \midrule
        
        \multirow{3}{*}{DE-NER} & \multirow{3}{*}{2} & 0 & 72.86$\pm$0.22 & 63.08$\pm$2.39 & 93.69$\pm$0.78 & \textbf{49.59$\pm$3.87} & 73.96$\pm$0.92 \\
         &  & 2 & \textbf{73.70$\pm$0.75} & 62.27$\pm$2.86 & 92.48$\pm$1.31 & 39.66$\pm$4.35 & 73.24$\pm$0.36 \\
         &  & 5 & 72.97$\pm$0.43 & 63.62$\pm$2.32 & 92.94$\pm$1.17 & 38.29$\pm$6.49 & 73.41$\pm$1.38 \\
        \hdashline

        \multirow{3}{*}{DE-NER} & \multirow{3}{*}{3} & 0 & 72.55$\pm$0.26 & 64.11$\pm$2.33 & 93.73$\pm$0.52 & 44.82$\pm$8.04 & 73.85$\pm$0.99 \\
         &  & 2 & 73.15$\pm$2.69 & 62.38$\pm$2.63 & 92.61$\pm$1.22 & 42.01$\pm$6.69 & 73.39$\pm$1.78 \\
         &  & 5 & 72.64$\pm$1.30 & 64.01$\pm$2.32 & 92.52$\pm$1.57 & 33.68$\pm$8.18 & 72.96$\pm$2.46 \\
        \hdashline

        \multirow{3}{*}{DE-NER} & \multirow{3}{*}{4} & 0 & 71.87$\pm$0.16 & \textbf{65.08$\pm$2.27} & \textbf{94.20$\pm$0.41} & 49.12$\pm$0.81 & \textbf{74.35$\pm$0.39} \\
         &  & 2 & 72.66$\pm$0.90 & 62.36$\pm$1.60 & 93.08$\pm$1.30 & 37.32$\pm$2.26 & 72.97$\pm$1.23 \\
         &  & 5 & 73.34$\pm$1.02 & 62.21$\pm$1.12 & 92.86$\pm$0.70 & 33.17$\pm$4.05 & 72.73$\pm$0.80 \\
        \bottomrule
    \end{tabular}
    }
    \caption{Results(\%) on CoNLL03 of DE-NER based on GPT-3.5-Turbo with numbers of demonstrations. The best scores are \textbf{bold}.}
    \label{tab:conll03_de_results}
\end{table}

\paragraph{Entity Type}
Table \ref{tab:overall_results} shows that DE-NER is beneficial for entities with high ambiguity, such as the MISC in CoNLL03 and WikiGold, with improvements of 27.95 scores and 13.75 scores, respectively. Additionally, for domain-specific entity, DE-NER achieves notable improvements on DNA (+7.42), Cell\_type (+1.96), and Protein (+4.36). 



\subsection{Analysis and Discuss}


\paragraph{Scaling Law}
We base Roleplayer and Responser to LLMs with different sizes and keep Questioner unchanged to investigate the effect of scaling law on our models. As shown in Table~\ref{tab:mistral_results}, the performance of the DE-NER based on smaller language model (Mistral-7B-instruct-v0.2) significantly decreases, while the performance increases if the model based on larger language model (GPT-4o-mini), suggesting that our models follow and are heavily affected by the scaling law. We conclude that although the learnable questioner is small, the performance of our models can be enhanced as the knowledge of theEMNLP frozen Roleplayer and Responser increases.  




\paragraph{In-Context Demonstrations}
We equip our model with several demonstrations for $k$-shot ($k \in \{2,5\}$) NER by in-context learning. The demonstrations are randomly sampled from the high-quality dialogue trajectories $\mathcal{D}^*_i$ selected in Section~\ref{generation}. As shown in Table~\ref{tab:conll03_de_results}, demonstrations cannot improve DE-NER. As the number of demonstration increases, the performance decreases consistently. The main reason is that DE-NER explicitly trained to perform instance-specific reasoning is limited to static in-context demonstrations of encouraging the model to focus on fixed surface features derived from a small set of examples. We conclude that LLMs know what LLMs need to accomplish tasks, which could be lagged by manual interferences.



\section{Conclusion}
We propose a dialogue elicitation of LLMs framework for zero-shot Named Entity Recognition that elicit large language models to correctly generate responses about named entities. The proposed framework consisting of a learnable questioner, a frozen roleplayer and a frozen responser self-plays to effectively resolves latent ambiguity of named entity recognition without requiring annotated data or manual demonstrations. Additionally, we propose the training with dialogue trajectories to make the model to ask clarifying questions that can effectively elicit LLMs to output correct answers. 
Experiments carried out on standard benchmarks demonstrate that our proposed model consistently outperforms the competitive baselines.

\section*{Limitations}
Despite the promising results achieved by our dialogue-based interaction framework, several limitations remain to be addressed in future work:
\begin{itemize}

\item This work focuses on zero-shot named entity recognition, and the proposed DE-NER has not yet been evaluated on other information extraction tasks or more diverse domains.

\item Our experiments are mainly conducted with relatively strong large language models, and the effectiveness of the proposed method on smaller models remains an open question.
\end{itemize}

\bibliography{acl_latex_short}
\newpage
\appendix
\section{Prompt Templates}
\label{appendix}
We show the prompts use in this work in Table \ref{tab:prompt_questioner} , Table \ref{tab:prompt_answerer} and Tabel \ref{tab:prompt_responser}.

\begin{table*}[!tp]
    \centering
    \small
    \renewcommand{\arraystretch}{1.3}
    \begin{tabular}{p{0.95\linewidth}}
        \toprule
        \textbf{Prompts of Questioner model} \\
        \midrule
        You are an expert in Named Entity Recognition (NER). \\
        Review the dialogue history. Then generate exactly \textbf{ONE} clarifying question that will most improve the accuracy of the NER result. \\
        \\
        Text: \{\} \\
        Preliminary NER result: \{\} \\
        Dialogue History: \{\} \\
        \\
        \textbf{Requirements:} \\
        - Output only one question. \\
        - Do not repeat questions in the dialogue history. \\
        - Your question should directly address one of the following issues in the preliminary NER result: \\
        \quad $\bullet$ [Entity Type Ambiguity]: If an entity's type is uncertain. \\
        \quad $\bullet$ \textbf[Entity Boundary Conflict]: If the span of an entity is unclear. \\
        \quad $\bullet$ \textbf[Missed Entity]: If a potential entity might have been omitted. \\
        \quad $\bullet$ \textbf[Entity Type Error]: If an entity might be mislabeled. \\
        \bottomrule
    \end{tabular}
    \caption{Prompt template for the Questioner model.}
    \label{tab:prompt_questioner}
\end{table*}

\begin{table*}[h!]
    \centering
    \begin{tabular}{p{0.95\linewidth}}
        \toprule
        \textbf{Prompts of Roleplayer model} \\
        \midrule
        You are an expert in Named Entity Recognition (NER). \\
        Dialogue History:{\{\}} \\
        Your task is review the dialogue history, answer the question:{\{\}} \\
        \bottomrule
    \end{tabular}
    \caption{Prompt template for the Roleplayer model.}
    \label{tab:prompt_answerer}
\end{table*}

\begin{table*}[h!]
    \centering
    \begin{tabular}{p{0.95\linewidth}}
        \toprule
        \textbf{Prompts of Responser model} \\
        \midrule
        You are an expert in Named Entity Recognition (NER). \\
        You should refer to the dialogue history as contextual information to improve entity recognition accuracy.  \\
        \\
        Text: \{\} \\
        Preliminary NER result: \{\} \\
        Dialogue History: \{\} \\
        \\
        Given the entity label set is: ['DNA', 'RNA', 'protein','cell\_line','cell\_type']. Based on the given entity label set, according to the genia annotation guidelines, please recognize the named entities in the given text. \\
        Provide answer in the following JSON format: [{{'Entity Name': 'Entity Label'}}]. If there is no entity in the text, return the following empty list: [] \\
        \bottomrule
    \end{tabular}
    \caption{Prompt template for the Responser model. Entity sets from GENIA are displayed.}
    \label{tab:prompt_responser}
\end{table*}

\section{Comparison Baselines}
We compare DE-NER with several representative zero-shot NER methods. The comparison results are reported in Table~\ref{tab:baseline_comparison}.
\begin{itemize}
    \item \textbf{Prompt-based}
    This baseline directly prompts the LLM for zero-shot entity recognition without dialogue interaction or demonstrations.

    \item \textbf{Self-Improving}~\cite{xie-etal-2024-self}.
    This method constructs reliable self-annotated demonstrations from an unlabeled corpus via self-consistency.

    \item \textbf{CMAS}~\cite{wang2025cooperative}.
    This method uses a cooperative multi-agent framework with self-annotation, type-related feature extraction, demonstration discrimination, and final prediction. 

    \item \textbf{ReverseNER}~\cite{wang2024reversener}.
    This method builds an example library by reversing the NER process from entity definitions to labeled sentences.
\end{itemize}

\begin{table}[H]
    \centering
    \caption{Results(\%) on CoNLL03, WikiGold, and GENIA.}
    \resizebox{\columnwidth}{!}{%
    \begin{tabular}{lcccc}
    \toprule
    \textbf{Method} & \textbf{CoNLL03} & \textbf{WikiGold} & \textbf{GENIA} & \textbf{AVG.} \\
    \midrule
    Prompt-based & 70.58 & 71.55 & 47.65 & 63.26 \\
    Self-Improving & 74.51 & 73.98 & 52.06 & 66.85 \\
    CMAS & -- & 76.23 & 50.00 & -- \\
    ReverseNER (GPT-4o mini) & 77.78 & 78.45 & -- & -- \\
    \midrule
    \textbf{DE-NER (Ours)} & 74.35 & 74.97 & 51.70 & 67.01 \\
    \bottomrule
    \end{tabular}%
    }
    \label{tab:baseline_comparison}
\end{table}

\section{Results on Additional Benchmarks}
We further evaluate DE-NER on the MIT-Restaurant dataset to reduce the potential contamination risk of widely used NER benchmarks and to test the method. Table~\ref{tab:mit_restaurant_results} reports results with both closed-source and open-source LLMs as Roleplayer and Responser. 

\section{Case Study}
Table \ref{tab:di-ner-example} shows that the prediction on GENIA across different models. In the first example, prompt-based model recognize DCs as a cell\_type, but fail to detect Sp1 while DE-NER identity correctly Sp1 as protein via dialogue interaction with LLMs of raising a question of "What is the biological role of Sp1 in the given context?" to elicit LLMs to make corrections. In the second examples, prompt-based model  fails to detect any entity (TR and TREp) while DE-NER correctly both of them by clarifying the definition of named entity. In the third examples, DE-NER can correct the entity type via further clarify questions. These examples show two typical effects of dialogue elicitation: recovering missed entities, clarifying error entity types. The generated clarification questions guide the model to revisit the initial prediction and use its internal knowledge more effectively for zero-shot NER.

\begin{table*}[!t]
    \centering
    \caption{Results(\%) with closed-source LLMs and open-source LLMs on MIT-Restaurant dataset. The best scores are \textbf{bold}.}
    \renewcommand{\arraystretch}{1.1}
    \setlength{\tabcolsep}{3pt}

    \resizebox{\textwidth}{!}{
    \begin{tabular}{lcccccccccc}
        \toprule
        \textbf{Model} & $N$ & \textbf{AME.} & \textbf{CUI.} & \textbf{DISH} & \textbf{HOUR} & \textbf{LOC.} & \textbf{PRI.} & \textbf{RAT.} & \textbf{Name.} & \textbf{Overall} \\
        \midrule

        \multicolumn{11}{c}{\textbf{GPT-4o-mini}} \\
        \midrule
        Prompt & -- & 9.90 & 39.39 & 26.67 & 15.38 & 27.49 & 33.33 & 36.07 & 66.67 & 35.12 \\
        \midrule
        \multirow{3}{*}{Base}
            & 2 & 20.33$\pm$0.91 & 41.78$\pm$1.43 & 39.03$\pm$0.48 & 21.56$\pm$3.66 & 29.26$\pm$2.21 & 38.98$\pm$1.57 & \textbf{41.60$\pm$1.81} & 68.10$\pm$0.51 & 40.00$\pm$0.65 \\
            & 3 & 19.88$\pm$1.58 & 41.79$\pm$2.64 & 36.15$\pm$1.71 & \textbf{24.71$\pm$0.25} & 29.85$\pm$0.72 & 37.66$\pm$1.99 & 40.00$\pm$2.18 & 68.30$\pm$2.51 & 39.93$\pm$0.59 \\
            & 4 & \textbf{21.95$\pm$3.39} & 38.97$\pm$1.93 & 33.63$\pm$2.01 & 23.24$\pm$3.79 & 29.70$\pm$1.03 & \textbf{40.60$\pm$1.20} & 39.10$\pm$3.67 & 69.70$\pm$0.70 & 39.72$\pm$0.55 \\
        \hdashline
        \multirow{3}{*}{DE-NER}
            & 2 & 19.94$\pm$2.61 & \textbf{45.10$\pm$2.64} & \textbf{47.33$\pm$1.88} & 18.95$\pm$3.46 & 33.52$\pm$2.88 & 40.19$\pm$0.33 & 39.70$\pm$1.33 & 70.30$\pm$0.93 & \textbf{42.43$\pm$1.06} \\
            & 3 & 20.04$\pm$3.22 & 43.79$\pm$0.53 & 46.11$\pm$3.46 & 18.19$\pm$4.65 & 33.19$\pm$0.79 & 40.20$\pm$3.39 & 38.25$\pm$1.89 & \textbf{71.48$\pm$0.50} & 42.24$\pm$0.41 \\
            & 4 & 18.53$\pm$6.16 & 43.75$\pm$1.69 & 43.90$\pm$3.21 & 19.37$\pm$4.92 & \textbf{33.74$\pm$2.23} & 39.80$\pm$1.72 & 37.80$\pm$1.52 & 71.46$\pm$0.76 & 42.05$\pm$0.51 \\

        \midrule
        \multicolumn{11}{c}{\textbf{Qwen3-14B}} \\
        \midrule
        Prompt & -- & 21.24 & 66.67 & 43.18 & 18.87 & 30.91 & 44.12 & 36.07 & 83.24 & 47.85 \\
        \midrule
        \multirow{3}{*}{Base}
            & 2 & \textbf{45.43$\pm$2.95} & \textbf{71.99$\pm$0.69} & \textbf{63.94$\pm$2.89} & 32.55$\pm$0.99 & 36.20$\pm$1.10 & \textbf{60.51$\pm$0.44} & \textbf{47.29$\pm$0.40} & 70.07$\pm$0.36 & \textbf{54.29$\pm$0.44} \\
            & 3 & 43.71$\pm$4.72 & 70.86$\pm$0.07 & 63.23$\pm$1.92 & 32.71$\pm$4.26 & 35.45$\pm$0.59 & 59.92$\pm$2.92 & 44.75$\pm$2.32 & 71.53$\pm$1.82 & 53.79$\pm$1.20 \\
            & 4 & 42.01$\pm$2.62 & 71.25$\pm$1.38 & 63.43$\pm$2.81 & \textbf{33.92$\pm$5.17} & \textbf{37.14$\pm$1.63} & 58.69$\pm$4.39 & 45.54$\pm$0.80 & 70.81$\pm$1.74 & 53.90$\pm$1.49 \\
        \hdashline
        \multirow{3}{*}{DE-NER}
            & 2 & 44.77$\pm$1.94 & 69.85$\pm$1.18 & 62.39$\pm$0.57 & 27.06$\pm$1.67 & 35.78$\pm$0.85 & 57.85$\pm$1.90 & 39.18$\pm$2.72 & 75.53$\pm$1.09 & 53.81$\pm$0.69 \\
            & 3 & 45.04$\pm$2.66 & 69.56$\pm$1.93 & 62.39$\pm$0.57 & 27.15$\pm$2.84 & 35.65$\pm$0.89 & 57.26$\pm$0.92 & 39.20$\pm$0.33 & 75.64$\pm$0.60 & 53.67$\pm$0.63 \\
            & 4 & 44.14$\pm$1.22 & 70.13$\pm$0.85 & 62.73$\pm$0.96 & 25.80$\pm$2.14 & 36.19$\pm$0.86 & 57.26$\pm$0.92 & 38.38$\pm$1.75 & \textbf{75.81$\pm$0.81} & 53.76$\pm$0.08 \\
        \bottomrule
    \end{tabular}
    }
    \label{tab:mit_restaurant_results}
\end{table*}

\begin{table*}[!tp]
\centering
\small
\setlength{\extrarowheight}{2pt} 
\begin{tabularx}{\textwidth}{
>{\raggedright\arraybackslash}X
>{\raggedright\arraybackslash}X
>{\raggedright\arraybackslash}X
>{\raggedright\arraybackslash}X
}
\toprule
\textbf{Input Sentence} & \textbf{Baseline Prediction} & \textbf{Dialogue Interaction} & \textbf{DE-NER Prediction} \\
\midrule

However, \textbf{[DCs]}{\textsubscript{cell\_type}} lack \textbf{[Sp1]}{\textsubscript{cell\_type}}, which may explain the failure of HIV-1 to replicate in purified DCs. &
\{"DCs": "cell\_type"\} &
Q: What is the biological role of Sp1 in the given context? \newline
A: Sp1 is a transcription factor regulating gene expression. &
\{"DCs": "cell\_type"\}, \{"Sp1": "protein"\} \\

\midrule

At high concentration of NE, all of the \textbf{[TR]}{\textsubscript{protein}} bound to \textbf{[TREp]}{\textsubscript{DNA}} was more greatly retarded than in the absence of NE. &
[] &
Q: What specific compound is referred to as NE in the text? \newline
A: NE is norepinephrine, a neurotransmitter likely affecting TR binding to TREp. &
\{{"TR": "protein"\}, \{"TREp": "DNA"}\} \\

\midrule

Human immunodeficiency virus \textbf{[vpr product]}{\textsubscript{protein}} is a \textbf{[virion-associated regulatory protein]}{\textsubscript{protein}}. &
\{'Human immunodeficiency virus': 'protein'\}, \{'vpr product': 'protein'\}, \{'virion-associated regulatory protein': 'protein'\} &
Q: What is the type of "Human immunodeficiency virus" in the text? \newline
A: The entity Human immunodeficiency virus in the text is a virus, not a protein. &
\{{'vpr product': 'protein'\}, \{'virion-associated regulatory protein': 'protein'}\} \\

\bottomrule
\end{tabularx}

\caption{GENIA dataset examples show baseline NER prediction, dialogue interaction, and DE-NER prediction.}
\label{tab:di-ner-example}
\end{table*}

\end{document}